# Neural Architecture for Fast and Reliable Coagulation Assessment in Clinical Settings: Leveraging Thromboelastography

Yulu Wang[1], Ziqian Zeng[2], Jianjun Wu[2], Zhifeng Tang[†1]

[1]Zhejiang University

[2]Huzhou Institute of Zhejiang University

{yulu.wang, zeng155, tangzhifeng}@zju.edu.cn, wujj@hizju.org

## Abstract

In an ideal medical environment, real-time coagulation monitoring can enable early detection and prompt remediation of risks. However, traditional Thromboelastography (TEG), a widely employed diagnostic modality, can only provide such outputs after nearly 1 hour of measurement. The delay might lead to elevated mortality rates. These issues clearly point out one of the key challenges for medical AI development: Making reasonable predictions based on very small data sets and accounting for variation between different patient populations, a task where conventional deep learning methods typically perform poorly. We present Physiological State Reconstruction (PSR), a new algorithm specifically designed to take advantage of dynamic changes between individuals and to maximize useful information produced by small amounts of clinical data through mapping to reliable predictions and diagnosis. We develop MDFE to facilitate integration of varied temporal signals using multi-domain learning, and jointly learn high-level temporal interactions together with attentions via HLA; furthermore, the parameterized DAM we designed maintains the stability of the computed vital signs. PSR evaluates with 4 TEG-specialized data sets and establishes remarkable performance -- predictions of $R^2 > 0.98$ for coagulation traits and error reduction around half compared to the state-of-the-art methods, and halving the inferencing time too. Drift-aware learning suggests a new future, with potential uses well beyond thrombophilia discovery towards medical AI applications with data scarcity.

## Introduction

Extracting meaningful information from limited and diverse datasets is a common predicament in training prediction models, even more so for those applications where failure directly affects human lives (Suresh et al., 2018). This compounds the issue as real-time updating is required, and the differences between instances can be so significant that the failings of standard deep learning arise. Literature identified two main strategies: few-shot learning and domain adaptation. Few-shot learning (An et al., 2025; Huang et al., 2024b), provided the ability to generalize from limited samples but is difficult due to the high variance in sample quality and distribution; whereas domain adaptation (Wang et al., 2024a; Feng et al., 2023) is able to transfer knowledge from one dataset to another provided that stable distributions are used during training, however, stable distributions are rare in clinical scenarios with dynamic and patient-specific instances. Thus, limited data with real-time drift adaptation has not been addressed properly. Examples of physiological monitoring challenges are evident in thromboelastography (TEG), used for evaluating coagulation. As a method to diagnose coagulation status by observing the viscoelasticity of blood, this method is helpful for gaining a real-time view of clot formation events, thus aiding greatly in treatment decisions during surgical situations. Conventional TEG testing requires time to complete, with results that are too late to inform accurate critical care decisions. A typical conventional TEG test would take nearly an hour, and an additional 15-minute delay increases the risk of trauma patients dying by approximately 10% (Gayet-Ageron et al., 2018). Predictive models for TEG need to be constructed. In fact, it reflects a broader problem: most current AI methods extract much less useful knowledge given very few observations; they also struggle with shifting distributions caused by the need to adjust for different populations. All of these contribute to poor model performance when there are changes in individual cases (Liu and Hauskrecht, 2017).

In recent years, the new neural network architectures provide advanced mechanisms to deal with varied kinds of data. For example, the Transformer variant (Liu et al., 2024a; Nie et al., 2023) can capture long-range dependency, while Kolmogorov-Arnold Networks (KANs) (Huang et al., 2024a; Genet and Inzirillo, 2024; Liu et al., 2024b). exhibits excellent capability of approximating complex functions, however, these models assume a static underlying distribution and sufficient training data, which is unrealistic in dynamic real-world scenarios with limited training data. The theoretical gap lies in the insurmountable learn-theoretic requirements for accurate predictions in restricted environments. Traditional PAC-learning bounds are loose under severe sample complexity restrictions (Cohen-Addad et al., 2025), and drift detection methods require sufficient data for stable baselines, which can harm the performance of adaptive learning systems.

To surmount the challenges, we propose an innovative drift-aware learning paradigm—Physiological State Reconstruction (PSR). PSR presents a novel mathematical scheme that simultaneously achieves feature extraction, prediction, and adaptation in part-time series data, ensuring reliable inference in limited data scenarios compared to existing methods. PSR includes three key components:

- **Multi-Domain Feature Extraction (MDFE)**: A method to achieve the time-frequency domain decomposition while preserving the signal's original features.
- **Hierarchical Learning Architecture (HLA)**: An integration of KANs and attention mechanisms that enhances approximation for complex physiological functions while maintaining interpretability.
- **Dynamic Adaptation Module (DAM)**: A real-time adaptation mechanism that facilitates incremental learning from minimal observation.

PSR introduces drift-aware learning to address AI deployment challenges in data-scarce, safety-critical conditions, enabling real-time adaptation and contributing to: (1) A new paradigm for safety-critical AI that improves responsiveness to data changes; (2) The theoretical basis of making reliable forecasts to achieve efficient information mining; (3) Utilize few case records to deliver clinical-grade care.

## Related Works

**Multi-domain Integration in Time Series.** Multi-domain feature extraction and fusion are essential for improving time series forecasting. While single-domain methods like ARIMA identify stable trends, they miss nonlinear dynamics in complex datasets (Li et al., 2023). Researchers are now integrating time and frequency domains. For instance, CTFNet combines convolutional mapping with time-frequency decomposition, reducing forecasting error by 64.7% (Zhang et al., 2024). TFMSNet uses multi-scale processing for effective feature fusion across 70 datasets (Song et al., 2025). These techniques are crucial for capturing intricate physiological data features that PSR aims to utilize.

**Prior Work on TEG Modeling.** Fast coagulation assessment and effective TEG modeling must be real-time. The Biological Mechanism-Driven Model (BPTM), using blood protein concentration, predicts TEG output to understand plasma coagulation in emergencies, effectively describing the relationships between blood proteins and patient coagulation (Ghetmiri et al., 2024). Besides, the knowledge that not all models can retain all the biological processes was also gained via KAN-based models—TimeKAN (Huang et al., 2024a) and TKAN (Genet and Inzirillo, 2024), Multilayer Perceptrons-based models—TimeMixer (Wang et al., 2024b), DLinear (Zeng et al., 2023), and FreTS (Yi et al., 2023), Transformer-based models—iTransformer (Liu et al., 2024a) and PatchTST (Nie et al., 2023), they generally omit most biological conditions and only integrate the related time series characteristics.

**Concept Drift Handling Techniques.** Concept drift handling techniques highly challenge time series analysis in the context of health care, and most recent approaches focus on actively detecting drift and passively adapting models via shifting models according to changes in data (Liu et al., 2023), some active methods fuse manifold projection and statistical process control to obtain better results (Wang et al., 2023). A hybrid feature extraction algorithm can detect drift occurring in a stream dataset more quickly because of incremental learning (Yu et al., 2022). Incremental learning also can cause the models to fit different sets of data according to data variations, for in-stance, StreamWNN increases accuracy as data are gradually added (Melgar-Garcia et al., 2023), and OneNet decreases errors by more than 50% (Wen et al., 2023). The importance of adapting to drift and keeping the model's capability has come up again. From PSR's point of view, it advocates to increase the DAM.

## Preliminaries

This section presents the Physiological State Reconstruction (PSR) framework, composed of two stages—the offline pretraining stage and the online prediction stage—which is provided with three main modules to conduct estimation.

**Offline Pretraining & Online Prediction.** Let $L_{\text{pred}}$ represent the number of historical TEG data points used in the online stage for accurate output estimations. During offline, pretraining the model scans the historical TEG series $X(t_1), ..., X(t_{L_{\text{pre}}})$ with an $N$-point sliding window $X_N(i) = [X(t_i), ..., X(t_{i+N-1})]$ and its timestamps $T_N(i) = [t_i, ..., t_{i+N-1}]$. The feature extractor converts each window into a matrix $X_{\text{multi}}(i) = \text{MDFE}(X_N(i), T_N(i)) \in \mathbb{R}^{N \times F_0}$, capturing salient time and frequency domain cues, with $F_0$ representing the total features derived from TEG data. The Hierarchical Learning Architecture (HLA) predictor $f_\theta^{\text{pre}}$, combined with the Dynamic Adaptation Module (DAM), is trained on these features, and its worst observed error

$$\delta_0 = \sup_{i \leq L_{\text{pre}}} \left| f_\theta^{\text{pre}}(X_{\text{multi}}(i)) - y(i) \right| \geq 0 \quad (1)$$

serves as the baseline for later online updates. After being pretrained, the model moves on to the online prediction process. A TEG curve $\hat{X} \in \mathbb{R}^M$ needs to be reconstructed given a partial sequence of recorded TEG traces $X = [X(t_1), ..., X(t_m)] \in \mathbb{R}^m$, where $M > m$.

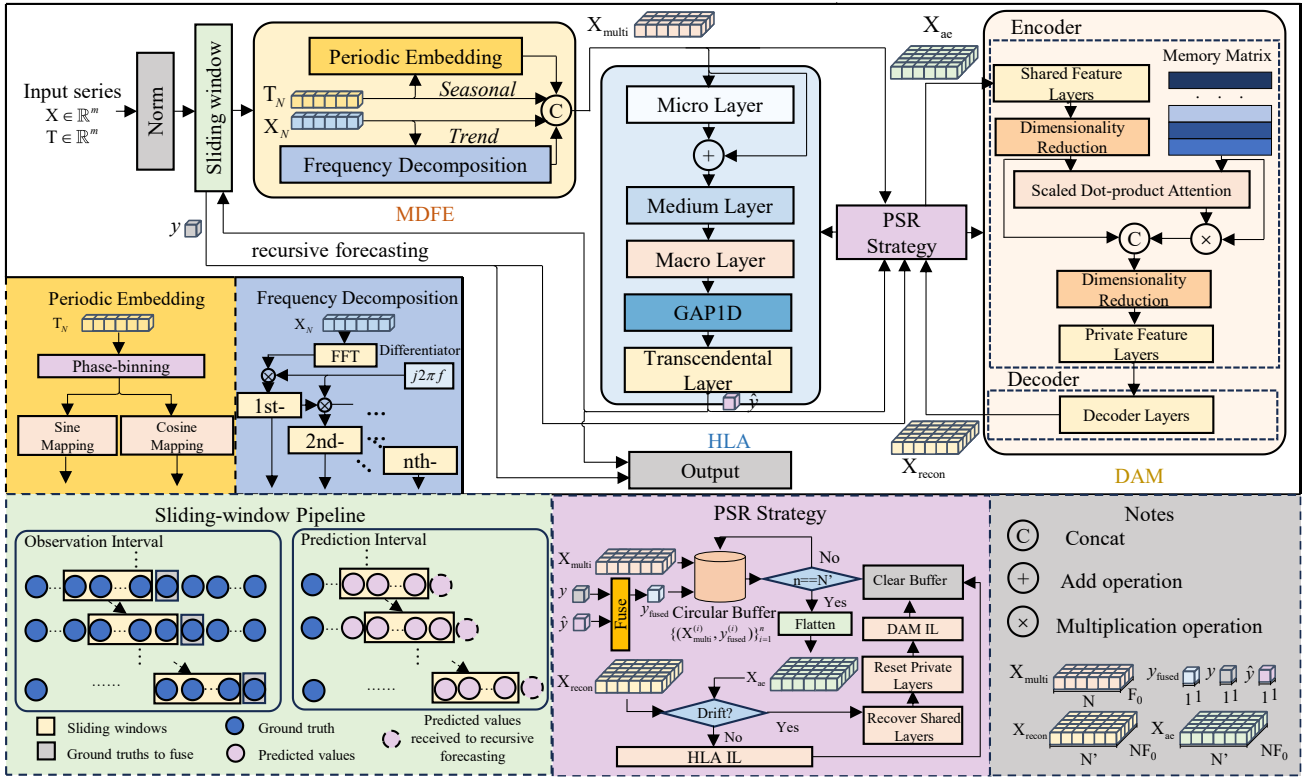

Figure 1: Overall structure of PSR. This figure illustrates the PSR framework. MDFE extracts key features across time and frequency domains. HLA combines Kolmogorov-Arnold networks with attention mechanisms for improved accuracy. DAM enables real-time learning from minimal data while preventing catastrophic forgetting.

The Index sets are: $I_{obs} = \{1, ..., m\}$, $I_{pred} = \{m+1, ..., M\}$, where $\beta(i) \in (0,1)$ smoothly increases across $I_{obs}$. The predictor is initialized with $f_\theta^{online} \leftarrow f_\theta^{pre}$. For each $i = 1, ..., M$:

**Sliding-Window & Feature Extraction.** Over the data, create sliding windows of $N$ $T_N(i) = [t_i, ..., t_{i+N-1}] \in \mathbb{R}^N$, $X_N(i) = [X(t_i), ..., X(t_{i+N-1})] \in \mathbb{R}^N$. Multi-channel features are obtained: $X_{multi}(i) = \text{MDFE}(X_N(i), T_N(i)) \in \mathbb{R}^{N \times F_0}$, $\hat{y}(i) = f_\theta^{online}(X_{multi}(i)) \in \mathbb{R}$.

**Fusion Target.** The fusion target should be defined:

$$y_{fused}(i) = \begin{cases} \beta(i) \cdot \hat{y}(i) + (1 - \beta(i)) \cdot y(i), & \text{if } i + N \in I_{obs}, \\ \hat{y}(i), & \text{otherwise}. \end{cases} \quad (2)$$

The fusion stage aligns predictions with observed values, so that estimates better represent reality.

**Circular Buffer & Recursion.** Store $(X_{multi}(i), y_{fused}(i))$ in a circular buffer of size. Only use $y_{fused}(i)$ for recursive forecasting if there are not enough entries less than $N'$.

**Concept Drift Detection.** Upon the buffer being filled up, flatten the features: $X_{adapt} \in \mathbb{R}^{N' \times (N \cdot F_0)}$. DAM reconstruction yields $X_{recon}$. The reconstruction loss is: $\mathcal{L}_{rec} = \|X_{adapt} - X_{recon}\|_2^2 / N'NF_0$. If the loss exceeds the threshold $\varepsilon$, that implies concept drift has occurred in the model. We need to reset DAM's shared layers and initialize its private layers currently; otherwise, only updating HLA. Thresholds are obtained analytically.

**Incremental Learning Gains & Error Sequence.** The incremental learning capabilities are evaluated through:

$$g_\theta(X) = f_\theta(\text{MDFE}(X, T)),$$
$$\gamma_i = \sup_X |g_{\theta_{i-1}}(X) - y_{fused}(i)| - \sup_X |g_{\theta_i}(X) - y_{fused}(i)| \geq 0, \quad (3)$$
$$\lambda_i = \sup_X |g_{\theta_i}(X) - y_{fused}(i)| - \sup_X |g_{\theta_i^+}(X) - y_{fused}(i)| \geq 0,$$
$$\delta_i = \delta_{i-1} - (\gamma_i + \lambda_i) \geq 0.$$

With $\gamma_i, \lambda_i \geq 0$, the sequence $\{\delta_i\}$ is nonincreasing, ensuring $\delta_i \geq 0$. Assuming $g_\theta(X)$ is $L$-Lipschitz in $X$.

**Theorem 1.** *(Adaptive Learning Convergence)*
*Under established assumptions, for any $1 \leq n \leq M$:*

$$\sum_{i=1}^n |\hat{y}(i) - y(i)| \leq \sum_{i=1}^n [1 + \beta(i)] L [\delta_0 - \sum_{j=1}^{i-1}(\gamma_j + \lambda_j)]. \quad (4)$$

This theorem establishes error bounds for our framework, ensuring manageable prediction errors as data grows. Nonnegative terms indicate model improvement with each update. The proof of Equation (4) is in Appendix A. A tighter $\mathcal{O}(\sqrt{T})$ *regret* result (proved in Appendix C.3) shows that the cumulative absolute error satisfies

$$\sum_{t=1}^T |\hat{y}_t - y_t| \leq (\delta_0 + G)\sqrt{T}. \quad (5)$$

Using adaptive steps and memory terms from DAM, our bound tightens Equation 4's $\mathcal{O}(T)$ to $\mathcal{O}(\sqrt{T})$, explaining why PSR oscillates back to steady state quickly even during drift. The TEG curve was reconstructed $\hat{X} = [[X(t_i)]_{i \in I_{obs}}, [\hat{y}(i)]_{i \in I_{pred}}] \in \mathbb{R}^M$.

# Methodology

## Multi-Domain Feature Extraction

The Multi-Domain Feature Extraction (MDFE) module is vital to the Physiological State Reconstruction (PSR) framework for processing TEG data using two operators for feature extraction.

**Periodic Embedding Operator ($P$)**: This operator is applied to capture the biological rhythm (seasonal features) with phase-binning and cycle length $C_j$ and period constants $P_j$ for $j = 1, ..., l$: $V_j(i) = \lfloor T_N(i) / P_j \rfloor \mod C_j \in \mathbb{Z}^N$. The corresponding periodic matrices are constructed as: $A_j(i) = [\sin(2\pi V_j(i)/C_j), \cos(2\pi V_j(i)/C_j)] \in \mathbb{R}^{N \times 2}$. Then we use the constructed periodic matrix to concatenate them and build the periodic feature matrix: $A(i) = [A_1(i), ..., A_l(i)] \in \mathbb{R}^{N \times 2l}$, which can refine the evaluation result on rhythms of physiological change to represent the coagulation status.

**Frequency Decomposition Operator ($F$)**: The operator decomposes a signal into the frequency domain using Fast Fourier Transform: $F_N(i) = \text{FFT}(X_N(i)) \in \mathbb{C}^N$. For each frequency bin $f_n = n/N$ and derivative order $k = 0, ... K-1$, features are calculated: $b_{k,n}(i) = (j2\pi f_n)^k F_N(i)[n]$. These coefficients reflect trend features, the real parts of them are preserved: $B(i) \in \mathbb{R}^{N \times K}$, $B(i)_{n,k} = \Re[b_{k,n}(i)]$.

The outputs from each operator create the feature set: $X_{\text{multi}}(i) = [T_N(i), X_N(i), A(i), B(i)] \in \mathbb{R}^{N \times F_0}$, where $F_0 = 2 + 2l + K$. This process captures both seasonal and trend patterns.

## Hierarchical Learning Architecture

The HLA enhances physiological signal prediction using a structured layered approach with input $X_{\text{multi}}(i) \in \mathbb{R}^{N \times F_0}$.

**Micro Layer (Mi-L)**: Use a Residual Convolutional Neural Network (ResCNN) to extract key local features for coagulation detection through convolutional patterns.

**Medium Layer (Me-L)**: Implement an LSTM network that captures temporal dependence. Thus, the model understands the past and can detect trends.

**Macro Layer (Ma-L)**: Employ a multi-head self-attention mechanism to identify global relationships among features, dynamically weighing their significance.

**Transcendental Layer (TL)**: Extract knowledge from previous layers using KANs to generate precise single-step prediction $\hat{Y}(i) \in \mathbb{R}$, integrating local and global context. The functional model:

$$\hat{Y}(i) = (\text{TL} \circ \text{Ma-L} \circ \text{Me-L} \circ \text{Mi-L}) X_{\text{multi}}(i) \quad (6)$$

**Theorem 2** *Let $D \subset \mathbb{R}^{N \times F_0}$ be a compact set and $f : D \to \mathbb{R}$ a continuous function. Denote by $\|\cdot\|$ the Euclidean norm on the ambient vector space. Then for every $\varepsilon > 0$ there exist integers $F, d, h, u, K$ and a choice of all trainable parameters $\theta$ in the HLA such that, for the model mapping $\hat{Y}(X_{\text{multi}}, \theta) : D \to \mathbb{R}$, we have:*

$$\sup_{X_{\text{multi}} \in D} \left| \hat{Y}(X_{\text{multi}}, \theta) - f(X_{\text{multi}}) \right| < \varepsilon \quad (7)$$

Theorem 2 ensures that with the right parameters, the model can approximate any continuous function over $D$ with bounded error and obtain the desired accuracy $\varepsilon$ which enables HLA to grasp complex information relationships for predicting results with high accuracy (Refer to Appendix A).

## Dynamic Adaptation Module

The Dynamic Adaptation Module (DAM) is a key component of the PSR framework, adapting to changes in physiological signals through three core insights introduced:

**1. Physiological Equilibrium Hypothesis**: Healthy individuals maintain physiological balance:

$$S = \{s \mid \|\Phi(s) - \mu\| < \varepsilon\}, \quad (8)$$

where $\Phi(s)$ is the physiological state and $\mu$ is the equilibrium reference.

**2. Pathological Drift Theorem**: Both pathological conditions and normal variations among individuals lead to non-linear shifts in physiological parameters:

$$D(t) = KL(P(t) \| P_{\text{baseline}}) > \text{threshold} \quad (9)$$

where $P(t)$ is the current parameter distribution and $P_{\text{baseline}}$ is the normative profile.

**3. Adaptive Convergence Principle:** The DAM features a memory-enhanced learning mechanism that improves adaptability to physiological signal changes by retaining relevant historical information for flexible adaptation to new data, used for re-balancing in the DAM:

$$\theta(t+1) = \theta(t) + \eta \nabla L + \beta \nabla M \quad (10)$$

Here, $\eta$ is the learning rate, $\nabla L$ is the loss gradient, and $\nabla M$ accounts for information from earlier times. More details and proof of these ideas can be found in Appendix C. Once a buffer is full, multiple channels of input get transformed to $X_{\text{adapt}} = \text{reshape}(X_{\text{multi}}, [N', N \cdot F_0]) \in \mathbb{R}^{N' \times (N \cdot F_0)}$. The DAM uses 3 shared KAN layers to extract features: $X_{\text{shared}} = (\text{KAN}_{\text{shared}}^{(3)} \circ \text{KAN}_{\text{shared}}^{(2)} \circ \text{KAN}_{\text{shared}}^{(1)}) X_{\text{adapt}} \in \mathbb{R}^{N' \times F_{\text{shared}}}$, where $F_{\text{shared}}$ specifies the feature dimension that matches basic physiological patterns.

These features are projected to a lower-dimensional space: $Q = X_{\text{shared}} \cdot W_{\text{proj}} + b_{\text{proj}} \in \mathbb{R}^{N' \times E}$, where $W_{\text{proj}}$ and $b_{\text{proj}}$ are reduction parameters. A learnable Memory Matrix (MM) captures historical data: $M_{\text{read}} = \text{softmax}(QM^T)M \in \mathbb{R}^{N' \times E}$. KAN maps inputs to 1-D features as keys, while memory $M$ holds recent target. Current and historical data are integrated as: $X_{\text{mem}} = [Q, M_{\text{read}}]W_{\text{mem}} + b_{\text{mem}} \in \mathbb{R}^{N' \times F_{\text{mem}}}$.

Then we use 3 private KAN layers to get $X_{\text{private}} \in \mathbb{R}^{N' \times F_{\text{private}}}$, where $F_{\text{private}}$ denotes the dimensionality of individual-specific features. The final output is generated through 3 decoder layers as $X_{\text{recon}} \in \mathbb{R}^{N' \times (N \cdot F_0)}$. The setup provides increased flexibility while keeping the model robust to the fluctuations in the individual's feature vectors.

# Experiments

## Simulation Study

This study is about simulating prediction models based on concept drift with the usage of synthetic data; it is shown that the proposed model can capture the variation from changes in the physiological signals.

We generated a synthetic dataset with predefined time series characteristics and introduced concept drift to mimic sudden transitions in human physiological states. As shown in Figure 2, the green line represents ground truth values, while the red dotted line denotes model predictions considering drift detection mechanisms.

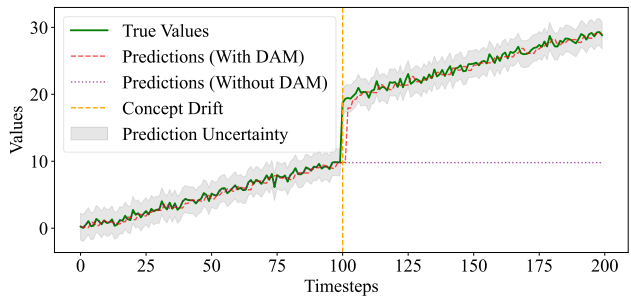

Figure 2: Simulation of Concept Drift Detection.

## Main Results

**Datasets.** We evaluated four TEG datasets for real-time coagulation curve reconstruction: HPP and HPC (healthy platelet-poor plasma), HWC (healthy whole blood), and TWA (trauma patient whole blood). These datasets encompass standard physiological conditions and the higher heterogeneity of trauma patients, ensuring robust model training and real-world applicability. Details are in Appendix B.

**Baselines.** We select 9 representative models to serve as baselines, including (1) Biological Mechanism-Driven model: BPTM (Blood Protein-Based TEG Model) (Ghetmiri et al., 2024); (2) KAN-based models: TimeKAN (Huang et al., 2024a), TKAN (Genet and Inzirillo, 2024), KAN (Liu et al., 2024b); (3) MLP-based models: TimeMixer (Wang et al., 2024b), DLinear (Zeng et al., 2023); (4) Transformer-based models: iTransformer (Liu et al., 2024a), PatchTST (Nie et al., 2023); (5) Frequency-based model: FreTS (Yi et al., 2023).

**Experimental Settings.** Five-fold cross-validation was adopted: 80% of the data were training and validation sets, and 20% was used as the test set. In terms of training, the 13th gen Intel® Core™ i9-13900HX CPU was utilized for GPU-free clinical real-time use. Every model applied Adam with an L2 loss function. More details are in Appendix D.

**Results.** Table 1 shows Mean Absolute Errors (MAE), Mean Squared Errors (MSE), and Coefficients of Determination ($R^2$) for each model on the HPP, HPC, HWC, and TWA, with the first place in bold font, the second place in underlined text, and the third place in dashed text. In all but one instance, PSR outperformed all the baselines by large margins, reducing MAE by 50–80%, and MSE by 45–95% relative to the best baseline (BPTM) while achieving $R^2$ scores >0.98. In TWA, BPTM only marginally outperforms PSR; however, both methods greatly exceed all others. Overall, PSR achieves an MAE of 0.056 and an MSE of 0.017, about 50% and 33% of BPTM's performances, showcasing exceptional real-time coagulation curve forecasting accuracy despite low sample sizes and non-stationary data.

## Ablation Study

To verify the effectiveness of each component of PSR, we provide a detailed ablation study on every possible design in MDFE, HLA, DAM, and PSR Strategy (Table 2).

**Study on MDFE.** In ablation ② (only the periodic embedding operator), removing the MDF component causes the MAE to increase from 0.056 to 0.383, and the R2 to decrease from 0.989 to 0.695. In ablation ③ (only frequency decomposition operator), removing TP embedding leads to an MAE of 0.482 and an R2 of 0.382. This shows that TP and MFD both play important roles in the coagulation model.

**Study on HLA.** Ablation ④–⑦ analyzed the contribution of every single layer:

- Ablation ④: Mi-L + TL (KAN).
- Ablation ⑤: Mi-L + Ma-L + TL (KAN).
- Ablation ⑥: Mi-Layer + Me-L + TL (KAN).
- Ablation ⑦: Mi-L + Me-L + Ma-L + TL (MLP).

This discovery underlines that almost every part is important for precise long-distance predictions, especially Me-L. Only the complete HLA achieved high performance, underscoring the need for integrating Mi-L, Me-L, Ma-L, and TL (KAN) for accurate coagulation reconstruction.

| Model | **PSR (Ours)** | | | BPTM (2024) | | | TimeKAN (2025) | | | TKAN (2024) | | | KAN (2024) | | | TimeMixer (2024) | | | DLinear (2023) | | | iTransformer (2024) | | | PatchTST (2023) | | | FreTS (2024) | | |
|---|---|---|---|---|---|---|---|---|---|---|---|---|---|---|---|---|---|---|---|---|---|---|---|---|---|---|---|---|---|---|
| Metric | MAE | MSE | $R^2$ | MAE | MSE | $R^2$ | MAE | MSE | $R^2$ | MAE | MSE | $R^2$ | MAE | MSE | $R^2$ | MAE | MSE | $R^2$ | MAE | MSE | $R^2$ | MAE | MSE | $R^2$ | MAE | MSE | $R^2$ | MAE | MSE | $R^2$ |
| HPP | **0.050** | **0.010** | **0.990** | 0.092 | 0.018 | 0.982 | 0.190 | 0.168 | 0.839 | 0.189 | 0.117 | 0.883 | 0.224 | 0.171 | 0.832 | 0.139 | 0.110 | 0.896 | 0.115 | 0.063 | 0.937 | 0.098 | 0.042 | 0.960 | 0.273 | 0.269 | 0.743 | 0.139 | 0.089 | 0.916 |
| HPC | **0.034** | **0.004** | **0.996** | 0.132 | 0.050 | 0.809 | 0.271 | 0.344 | 0.669 | 0.409 | 0.305 | 0.690 | 0.649 | 0.733 | 0.248 | 0.359 | 0.517 | 0.509 | 0.284 | 0.248 | 0.759 | 0.296 | 0.311 | 0.711 | 0.241 | 0.131 | 0.870 | 0.270 | 0.272 | 0.740 |
| HWC | **0.060** | **0.011** | **0.989** | 0.342 | 0.224 | 0.783 | 0.146 | 0.075 | 0.929 | 0.239 | 0.143 | 0.863 | 0.225 | 0.090 | 0.909 | 0.187 | 0.129 | 0.878 | 0.270 | 0.211 | 0.798 | 0.147 | 0.103 | 0.903 | 0.391 | 0.344 | 0.660 | 0.188 | 0.128 | 0.878 |
| TWA | 0.062 | 0.023 | 0.979 | **0.047** | **0.005** | **0.980** | 0.144 | 0.135 | 0.872 | 0.213 | 0.223 | 0.789 | 0.195 | 0.192 | 0.818 | 0.136 | 0.129 | 0.877 | 0.153 | 0.142 | 0.865 | 0.158 | 0.151 | 0.856 | 0.171 | 0.156 | 0.852 | 0.162 | 0.142 | 0.864 |
| Overall | **0.056** | **0.017** | **0.989** | 0.109 | 0.048 | 0.922 | 0.168 | 0.161 | 0.912 | 0.246 | 0.215 | 0.870 | 0.275 | 0.261 | 0.860 | 0.180 | 0.190 | 0.899 | 0.190 | 0.164 | 0.900 | 0.174 | 0.160 | 0.907 | 0.226 | 0.191 | 0.879 | 0.181 | 0.157 | 0.908 |

Table 1: Main results. The look-back window was set as 55% of the sequence length for HPP, but it was reduced to 45% in the case of longer datasets to minimize redundancy. Across the five folds, we report MAE, MSE and $R^2$ values.

| | MDFE | | HLA | | | | DAM | | PSR Strategy | | HPP | | | HPC | | | HWC | | | TWA | | | Overall | | |
|---|---|---|---|---|---|---|---|---|---|---|---|---|---|---|---|---|---|---|---|---|---|---|---|---|---|
| | P | F | Mi-L | Me-L | Ma-L | TL | MLP/KAN | MM | DAM | fusion | MAE | MSE | $R^2$ | MAE | MSE | $R^2$ | MAE | MSE | $R^2$ | MAE | MSE | $R^2$ | MAE | MSE | $R^2$ |
| ① | ✓ | ✓ | ✓ | ✓ | ✓ | KAN | KAN | ✓ | ✓ | ✓ | **0.050** | **0.010** | **0.990** | **0.034** | **0.004** | **0.996** | **0.060** | **0.011** | **0.989** | **0.062** | **0.023** | **0.979** | **0.056** | **0.017** | **0.989** |
| ② | ✓ | ✗ | ✓ | ✓ | ✓ | KAN | KAN | ✓ | ✓ | ✓ | 0.261 | 0.289 | 0.707 | 0.453 | 0.664 | 0.319 | 0.285 | 0.281 | 0.716 | 0.407 | 0.531 | 0.490 | 0.383 | 0.493 | 0.695 |
| ③ | ✗ | ✓ | ✓ | ✓ | ✓ | KAN | KAN | ✓ | ✓ | ✓ | 0.339 | 0.408 | 0.597 | 0.887 | 2.161 | -1.081 | 0.613 | 1.502 | -0.468 | 0.358 | 0.714 | 0.310 | 0.482 | 1.048 | 0.382 |
| ④ | ✓ | ✓ | ✓ | ✗ | ✗ | KAN | KAN | ✓ | ✓ | ✓ | 0.154 | 0.126 | 0.866 | 0.550 | 2.222 | -0.951 | 0.086 | 0.023 | 0.977 | - | - | - | 0.285 | 0.924 | 0.604 |
| ⑤ | ✓ | ✓ | ✓ | ✗ | ✓ | KAN | KAN | ✓ | ✓ | ✓ | 0.162 | 0.121 | 0.875 | 0.320 | 0.328 | 0.683 | 0.208 | 0.323 | 0.682 | - | - | - | 0.244 | 0.285 | 0.802 |
| ⑥ | ✓ | ✓ | ✓ | ✓ | ✗ | KAN | KAN | ✓ | ✓ | ✓ | 0.299 | 0.721 | 0.231 | 0.516 | 1.003 | 0.057 | 0.154 | 0.145 | 0.855 | 0.208 | 0.236 | 0.775 | 0.256 | 0.383 | 0.809 |
| ⑦ | ✓ | ✓ | ✓ | ✓ | ✓ | MLP | KAN | ✓ | ✓ | ✓ | 0.204 | 0.172 | 0.818 | 0.548 | 0.912 | 0.084 | 0.066 | 0.015 | 0.985 | 0.344 | 0.571 | 0.452 | 0.321 | 0.505 | 0.701 |
| ⑧ | ✓ | ✓ | ✓ | ✓ | ✓ | KAN | KAN | ✗ | ✓ | ✓ | 0.191 | 0.190 | 0.810 | 0.324 | 0.613 | 0.412 | 0.291 | 0.698 | 0.311 | 0.330 | 0.653 | 0.365 | 0.311 | 0.611 | 0.598 |
| ⑨ | ✓ | ✓ | ✓ | ✓ | ✓ | KAN | MLP | ✓ | ✓ | ✓ | 0.125 | 0.099 | 0.894 | 0.110 | 0.039 | 0.964 | 0.114 | 0.086 | 0.910 | 0.140 | 0.188 | 0.819 | 0.130 | 0.141 | 0.906 |
| ⑩ | ✓ | ✓ | ✓ | ✓ | ✓ | KAN | MLP | ✗ | ✓ | ✓ | 0.159 | 0.146 | 0.851 | 0.152 | 0.065 | 0.938 | 0.117 | 0.088 | 0.909 | 0.110 | 0.061 | 0.942 | 0.122 | 0.072 | 0.956 |
| ⑪ | ✓ | ✓ | ✓ | ✓ | ✓ | KAN | ✗ | ✗ | ✗ | ✓ | 0.477 | 0.911 | 0.081 | 0.428 | 0.526 | 0.458 | 0.518 | 0.720 | 0.286 | 0.929 | 1.977 | -0.906 | 0.747 | 1.459 | 0.030 |
| ⑫ | ✓ | ✓ | ✓ | ✓ | ✓ | KAN | KAN | ✓ | ✓ | ✗ | 0.172 | 0.222 | 0.783 | 0.121 | 0.049 | 0.951 | 0.156 | 0.187 | 0.816 | 0.211 | 0.379 | 0.644 | 0.185 | 0.283 | 0.815 |

Table 2: Ablation study results on each component of PSR. Ablations ④ and ⑤ did very poorly on the TWA dataset, leading to their absence from the table (denoted as "–"). Their scores were based on only three non-TWA datasets.

**Study on DAM.** Ablations confirm the KAN–memory synergy. Removing the memory matrix (MM) from KAN (⑧) wipes out its recall and collapses performance. Swapping KAN for an MLP (⑨, ⑩) shows that MLPs lack the read/write bias: with MM, they mostly fetch noise; without MM, they train stably but gain only minor generalisation. Only the full KAN + MM setup (①) excels—the Kolmogorov–Arnold keys let MM retrieve the right rows. Thus, as predicted in Appendix C.3, MM helps only with KAN, and neither component shines alone.

**Study on PSR Strategy.** The PSR study in ablation experiments. As seen in ablation ⑪, removal of the DAM leads to poor long-horizon performance; whereas, ablation ⑫, without the fusion mechanism but with DAM incorporated, only demonstrates a slight decrease in performance when compared to the standard model, implying that incremental update of DAMs and adaptively weight drift-aware fusion is required for robust and precise prediction.

## Model Analysis

**Varying Look-back Window.** Varying the size of the window from PSR (25%-55%). The corresponding results can be found in Table 3 for different datasets, including HPP, HPC, HWC, and TWA, under MAE, MSE, and $R^2$.

Expanding the look-back window improves accuracy by up to 45%, with diminishing returns beyond that. Increasing from 25% to 35% reduces MAE from 0.273 to 0.163 and raises $R^2$ from 0.800 to 0.907. A further increase to 45% decreases MAE to 0.072 and elevates $R^2$ above 0.98. Beyond 45%, gains taper off, indicating this window effectively captures core dynamics for accurate forecasting. The HPP series is poor at 45% ($R^2$ = 0.767) but improves at 55% ($R^2$ = 0.990), so we use 55% for HPP and 45% for HPC, HWC, and TWA. TWA obtained an $R^2$ of 0.925 and MAE of 0.127 for only 25%, which was due to trauma-induced inflection points, varying clotting response in different persons, and sharp biomarker transitions within the initial phase of data.

| Points Known Portion | | 25% | 35% | 45% | 55% |
|---|---|---|---|---|---|
| HPP | MAE | 0.595 | 0.512 | 0.256 | **0.050** |
| | MSE | 1.153 | 0.920 | 0.249 | **0.010** |
| | $R^2$ | -0.078 | 0.143 | 0.767 | **0.990** |
| HPC | MAE | 0.414 | 0.234 | 0.034 | **0.034** |
| | MSE | 0.686 | 0.325 | 0.004 | **0.003** |
| | $R^2$ | 0.296 | 0.686 | 0.996 | **0.997** |
| HWC | MAE | 0.515 | 0.293 | 0.060 | **0.058** |
| | MSE | 0.786 | 0.343 | 0.011 | **0.010** |
| | $R^2$ | 0.216 | 0.650 | 0.989 | **0.990** |
| TWA | MAE | 0.127 | 0.063 | 0.062 | **0.037** |
| | MSE | 0.080 | 0.024 | 0.023 | **0.006** |
| | $R^2$ | 0.925 | 0.978 | 0.979 | **0.994** |
| Overall | MAE | 0.273 | 0.163 | 0.072 | **0.041** |
| | MSE | 0.375 | 0.194 | 0.036 | **0.007** |
| | $R^2$ | 0.800 | 0.907 | 0.984 | **0.996** |

Table 3: A comprehensive performance comparison by varying look-back window.

**Model Efficiency.** Table 4 shows that the training process of the PSR takes 0.035s/0.26MB per step—slower than iTransformer (0.024s/0.85MB)—but still much smaller than TKAN (0.539s/25.5MB). For inference, it uses 0.041s/0.051MB, which is slower than ultralight methods like PatchTST ($5\times10^{-5}$s/0.0019MB) and DLinear ($2\times10^{-5}$s), but real time nevertheless. PSR alone features adaptive inference, however, it can take an extra 0.28s/5.05MB when using extra DAM updates during inference. Overall, it supports near real-time with top accuracy despite only moderate computational resources required.

**Model Interpretability.** SHAP quantified feature contributions for one-step clot strength predictions. For N-step windows, averaged 8 inputs (time, clot strength, sin/cos(time), FFT, and its 3 derivatives) into features. Computed on 100 random samples. Details in Appendix F.

| Models | Training Time (s/iter) | Training ΔRSS (MB/iter) | Pure Inference Time(s/step) | Pure Inference ΔRSS (MB/step) | Adaptive Inference Time (s/step) | Adaptive Inference ΔRSS (MB/step) |
|---|---|---|---|---|---|---|
| PSR | 0.035 | **0.26** | 4.10e-2 | 5.10e-2 | **0.28** | **5.05** |
| TimeKAN | 0.043 | 2.54 | 8.89e-5 | 5.74e-3 | - | - |
| TimeMixer | 0.039 | 1.44 | 6.97e-5 | 2.23e-3 | - | - |
| TKAN | 0.539 | 25.48 | 2.87e-3 | 8.60e-2 | - | - |
| KAN | 0.089 | 5.91 | 8.51e-4 | 1.47e-2 | - | - |
| iTransformer | 0.024 | 0.85 | 3.16e-5 | 2.65e-3 | - | - |
| FreTS | 0.019 | 7.93 | 5.70e-5 | 4.80e-3 | - | - |
| PatchTST | 0.010 | 0.71 | 5.06e-5 | **1.88e-3** | - | - |
| DLinear | **0.002** | 0.37 | **2.22e-5** | 1.90e-3 | - | - |

Table 4: Comparative analysis of model performance and inference efficiency. This table compares models based on their training times (seconds per iteration) and resource consumption. ΔRSS (Change in Resident Set Size) indicates the difference in memory usage (MB) during operations, reflecting additional memory utilized during training and inference.

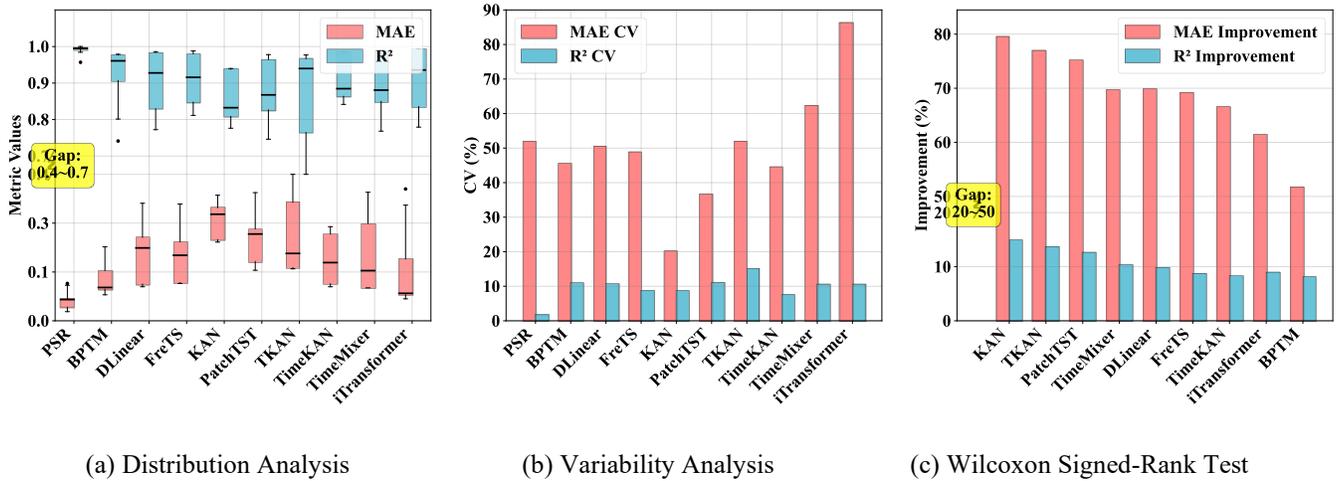

(a) Distribution Analysis  (b) Variability Analysis  (c) Wilcoxon Signed-Rank Test

Figure 3: Overall statistical evaluation of model performance.

## Statistical Robustness Analysis

To ensure reliability, we performed statistical analyses beyond mean comparisons, as shown in Figure 3, which includes distribution characteristics, performance stability, and significance testing using 5-fold cross-validation.

**Distribution Analysis (Figure 3a).** A Box plot shows different distributions of performance. PSR has a much better central tendency with the mini-mum of outlier appearance than competitors for both MAE and $R^2$, while the other methods exhibit wide distribution and more outlying values especially for larger errors.

**Variability Analysis (Figure 3b).** PSR showed impressive stability by calculating the coefficient of variation, CV=(σ/μ)×100%, where σ is the standard deviation, μ is mean. For PSR, the CV was 1% for $R^2$ and 52% for MAE.

**Statistical Significance (Figure 3c).** Wilcoxon signed-rank tests confirmed significant improvements for PSR across all comparisons, with MAE reductions of 51% to 79% and $R^2$ improvements of 8% to 15% compared to the best baselines, indicated by p-values<0.05.

**Note:** Due to the small sample size (n=5 folds), we employed non-parametric Wilcoxon signed-rank tests alongside coefficient of variation analysis to provide robust evidence of statistical and practical significance. All statistical tests were conducted with α = 0.05.

## Conclusion

This paper introduces Physiological State Reconstruction (PSR), a drift-aware framework of real-time coagulation assessment with minimum data. MDFE, HLA, and DAM come along as the basic parts of it. The combination of them forms a new medical AI algorithm. It has proven good effects with these evaluations: $R^2 > 0.98$, a decrease in MAE over 50%, and the diagnostic time being half the speed of traditional Thromboelastography. Moreover, it delivers up to sub-second inference speeds. Outside of coagulation assessments, the model is a pioneer in drift-aware learning as a mechanism that is formed for highly dynamic, data-scarce medical use cases, filling essential gaps for adaptive deployment of deep learning-based AI.

## Ethics Statement

This research adheres to established ethical guidelines for artificial intelligence and machine learning studies, including those outlined by the AAAI Code of Ethics and Professional Conduct. All datasets utilized in this work are publicly available from online sources, ensuring no involvement of human subjects, personal data collection, or experimental interventions that could raise privacy or consent concerns. Detailed descriptions of the datasets, including their sources, licenses, and any preprocessing steps, are provided in the Appendix to promote transparency and reproducibility.

We have assessed potential biases in the datasets. The study does not pose risks of harm to individuals or society, and we encourage responsible use of the proposed methods to avoid unintended application. No conflicts of interest exist, and all work complies with relevant data protection regulations, including open-source licensing terms.

## Acknowledgments

This work was supported by the National Natural Science Foundation of China (Grant No. 12474469) and the Huzhou Science and Technology Plan Project (No. 2024GZ15). We would also like to thank our colleagues for their valuable discussions and the reviewers for their constructive feedback.

# A Theoretical Proofs

## A.1 Proof of Theorem 1

**Theorem 1.** *(Adaptive Learning Convergence)*
Under these established assumptions, for any $1 \le n \le M$:

$$\sum_{i=1}^{n} |\hat{y}(i) - y(i)| \le \sum_{i=1}^{n} [1 + \beta(i)] L \Big[\delta_0 - \sum_{j=1}^{i-1} (\gamma_j + \lambda_j)\Big]. \quad (11)$$

**Proof of Theorem 1**

For each $i = 1, \ldots, n$, decompose the pointwise error:

$$|\hat{y}(i) - y(i)| \le A_i + B_i,$$
$$A_i = |\hat{y}(i) - y_{\text{fused}}(i)|, \quad (12)$$
$$B_i = |y_{\text{fused}}(i) - y(i)|.$$

**1. Control of $A_i$.**

Let $X_i^*$ be the 'ideal' window for true observation and $X_i$ be the actual window (which may contain past predictions). Then by Lipschitz continuity and the inductive bound $|g_{\theta_{i-1}}(X) - y_{\text{fused}}(i)| \le \delta_{i-1}$ for all $X$,

$$A_i = |g_{\theta_{i-1}}(X_i) - y_{\text{fused}}(i)| \le \underbrace{|g_{\theta_{i-1}}(X_i) - g_{\theta_{i-1}}(X_i^*)|}_{\le L\|X_i - X_i^*\| \le L\delta_{i-1}} + \underbrace{|g_{\theta_{i-1}}(X_i^*) - y_{\text{fused}}(i)|}_{\le \delta_{i-1}} \le L\delta_{i-1}. \quad (13)$$

**2. Control of $B_i$.**

By the fusion definition,

$$y_{\text{fused}}(i) = \begin{cases} \beta(i) \cdot \hat{y}(i) + (1 - \beta(i)) \cdot y(i), & \text{if } i + N \in I_{\text{obs}}, \\ \hat{y}(i), & \text{otherwise.} \end{cases} \quad (14)$$

One checks in both cases that

$$B_i \le \beta(i) |\hat{y}(i) - y(i)| \le \beta(i) L \delta_{i-1}. \quad (15)$$

Combing,

$$|\hat{y}(i) - y(i)| \le A_i + B_i \le [1 + \beta(i)] L \delta_{i-1} = [1 + \beta(i)] L \Big[\delta_0 - \sum_{j=1}^{i-1} (\gamma_j + \lambda_j)\Big]. \quad (16)$$

Summing over $i = 1, \ldots, n$ yields the stated bound.
Proved.

## A.2 Proof of Theorem 2

**Theorem 2** Let $D \subset \mathbb{R}^{N \times F_0}$ be a compact set and $f : D \to \mathbb{R}$ a continuous function. Denote by $\|\cdot\|$ the Euclidean norm on the ambient vector space. Then for every $\varepsilon > 0$ there exist integers $F, d, h, u, K$ and a choice of all trainable parameters $\theta$ in the HLA model such that, for the model mapping $\hat{Y}(X_{\text{multi}}, \theta) : D \to \mathbb{R}$, we have:

$$\sup_{X_{\text{multi}} \in D} |\hat{Y}(X_{\text{multi}}, \theta) - f(X_{\text{multi}})| < \varepsilon \quad (17)$$

**Proof of Theorem 2**

The definition of $F, d, h, u, K$ and mapping process of parameters can be found in the **Model Mapping Description**.

**1. Uniform continuity of $f$.**

Since $f$ is continuous on the compact set $D$, it is uniformly continuous. Thus there exists $\delta > 0$ such that

$$\|X - Y\| < \delta \Rightarrow |f(X) - f(Y)| < \frac{\varepsilon}{2}, \quad (18)$$

for all $X, Y \in D$.

**2. Encoder approximates an isometric embedding.**

Flatten each $X_{\text{multi}}$ into a vector in $\mathbb{R}^n, n = NF_0$. Fix any embedding dimension $d \ge n$ and define the linear embedding

$$L : \mathbb{R}^n \to \mathbb{R}^d, \quad L(x) = (x, 0) \in \mathbb{R}^d. \quad (19)$$

Choose the encoder parameters such that: $\theta_{\text{enc}} = \{W_1, W_2, W_r, \ldots\}$ so that:

• Each convolutional layer and each skip connection sets to zero all components except the first n ones, which stay the same.
• LSTM gates are set to keep hidden and cell states the same.
• Multi-head attention projects very close to identity.
• Global pooling is incredibly simple (e. g., replicating each coordinate across all time steps).

We can ensure:

$$\sup_{X_{\text{multi}} \in D} \|E_{\theta_{\text{enc}}} - L(\text{vec}(X_{\text{multi}}))\| < \delta \quad (20)$$

**3. KAN approximates $f \circ L^{-1}$.**

Define:

$$g : L(D) \subset \mathbb{R}^d \to \mathbb{R}, \quad g(z) = f(L^{-1}(z)). \quad (21)$$

By the Kolmogorov–Arnold theorem, for any $\eta > 0$ there exist univariate functions $\{\Phi_j, \varphi_{j,m}\}$ and integer $u$ such that:

$$\sup_{z \in L(D)} \Big| g(z) - \sum_{j=1}^{u} \Phi_j \Big( \sum_{m=1}^{d} \varphi_{j,m}(z_m) \Big) \Big| < \eta. \quad (22)$$

Moreover, each continuous $\Phi_j, \varphi_{j,m}$ on a compact domain can be uniformly approximated by a linear combination of SiLU plus $K$ B-spline basis functions. Consequently, by choosing $K$ and the KAN parameters $\theta_{\text{KAN}}$ sufficiently large, we obtain:

$$\sup_{z \in L(D)} |A_{\theta_{\text{KAN}}}(z) - g(z)| < \frac{\varepsilon}{2}. \quad (23)$$

**4. Composition and error bound.**

Define the full model:

$$\hat{Y}(X_{\text{multi}}; \theta) = A_{\theta_{\text{KAN}}}(E_{\theta_{\text{KAN}}}(X_{\text{multi}})). \quad (24)$$

Then for every $X_{\text{multi}} \in D$, let $z = E_{\theta_{\text{enc}}}(X_{\text{multi}})$. We have:

$$|\hat{Y}(X_{\text{multi}}; \theta) - f(X_{\text{multi}})| \le |A_{\theta_{\text{KAN}}}(z) - g(z)| + |g(z) - f(X_{\text{multi}})| \quad (25)$$

The first term is $< \frac{\varepsilon}{2}$ by step 3. Then the second term is

$$|f(L^{-1}(z)) - f(X_{multi})| = |f(L^{-1}(E_{\theta_{enc}}(X_{multi}))) - f(X_{multi})| < \frac{\varepsilon}{2}, \quad (26)$$

because $\|E_{\theta_{enc}} - L(\text{vec}(X_{multi}))\| < \delta$ and $f$ is uniformly continuous. Hence $|\hat{Y} - f| < \varepsilon$ uniformly on $D$.

Proved.

**Model Mapping Description**

We decompose the mapping $X_{multi} \mapsto \hat{Y}$ into two modules:

$$X_{multi} \xrightarrow{\text{Encoder } E_{\theta_{enc}}} O_{pool} \xrightarrow{\text{KAN } A_{\theta_{KAN}}} \hat{Y}. \quad (27)$$

**1. Residual Convolution Block (Micro Layer)**

Input $X_{multi} \in \mathbb{R}^{N \times F_0}$ passes through two 1D convolutions of kernels $W_1 \in \mathbb{R}^{k \times F_0 \times F}, W_2 \in \mathbb{R}^{k \times F \times F}$ plus a parallel $1 \times 1$ skip convolution $W_r \in \mathbb{R}^{1 \times F_0 \times F}$. Writing "*" for same-padding 1D conv:

$$\begin{aligned} R &= X_{multi} * W_r + b_r, \\ S &= (X_{multi} * W_1 + b_1) * W_2 + b_2, \\ H &= R + S \in \mathbb{R}^{N \times F}, \end{aligned} \quad (28)$$

yielding $H \in \mathbb{R}^{N \times F}$. Here $b_1$, $b_2$ and $b_r$ are learnable bias vectors in $\mathbb{R}^F$, added channelwise and broadcast across time and batch dimensions.

**2. LSTM Module (Medium Layer)**

Treat each row $H^{(t)} \in \mathbb{R}^F$ as timestep input $x_t$. With hidden size $d$, gates $f_t, i_t, o_t$ and cell update $\tilde{c}_t$ as usual, we obtain hidden states $h_t \in \mathbb{R}^d$. Stacking:

$$L = [h_1, \ldots, h_N] \in \mathbb{R}^{N \times d}, \quad (29)$$

**3. Multi-head Self-attention Module (Macro Layer)**

Given $L \in \mathbb{R}^{N \times d}$, we first compute

$$\begin{aligned} Q &= LW_Q + b_Q, K = LW_K + b_K, V = LW_V + b_V, \\ &(W_{\{Q,K,V\}} \in \mathbb{R}^{d \times d}, b_{\{Q,K,V\}} \in \mathbb{R}^d), \end{aligned} \quad (30)$$

then partition the feature axis into $h$ blocks of size $d/h$ via

$$S_h : \mathbb{R}^{N \times d} \to \mathbb{R}^{N \times h \times (d/h)}, (S_h(X))_{t,k,m} = X_{t,(k-1) \times \frac{d}{h} + m}, \quad (31)$$

and permute to bring the head dimension forward:

$$\wp : \mathbb{R}^{N \times h \times (d/h)} \to \mathbb{R}^{h \times N \times (d/h)}, \wp(X)_{k,t,m} = X_{t,k,m}. \quad (32)$$

This yields $\hat{Q}, \hat{K}, \hat{V} \in \mathbb{R}^{h \times N \times (d/h)}$. For each head $k$,

$$\begin{aligned} A^{(k)} &= \text{softmax}(\hat{Q}^{(k)} (\hat{K}^{(k)})^T / \sqrt{d/h}) \in \mathbb{R}^{N \times N}, \\ Z^{(k)} &= A^{(k)} \hat{V}^{(k)} \in \mathbb{R}^{N \times (d/h)}. \end{aligned} \quad (33)$$

Concatenating $\{Z^{(k)}\}_{k=1}^h$ and project to obtain $O \in \mathbb{R}^{N \times d}$.

**4. Global Average Pooling**

Averaging $O$ over time to obtain

$$O_{pool} \in \mathbb{R}^d, O_{pool}^{(m)} = \frac{1}{N} \sum_{t=1}^{N} O^{(t,m)}. \quad (34)$$

**5. Two-Layer Kolmogorov–Arnold Network (Transcendental Layer)**

Inspired by the Kolmogorov–Arnold representation theorem, which guarantees that any continuous mapping $f : \mathbb{R}^n \to \mathbb{R}$ can be written as

$$f(x_1, x_2, \ldots, x_n) = \sum_{i=1}^{m} \Phi_i \left( \sum_{i=1}^{n} \varphi_{ij}(x_j) \right), \quad (35)$$

we use a cascade of univariate splines and an affine map as the outer transformation for every KAN layer.

Each inner basis function $\varphi_{j,m}^{(\ell)}(x)$ ($\ell = 1, 2$, unit $j$, input index $m$) can be described as:

$$\varphi_{j,m}^{(\ell)}(x) = \text{SiLU}(x) + \sum_{k=1}^{K} \omega_{j,m,k}^{(\ell)} B_{k,p}(x), \quad (36)$$

where $\text{SiLU}(x) = x/(1+e^{-x})$ provides a smooth global nonlinearity. The integer $K$ denotes the total number of B-spline basis functions—namely, the number of knot intervals plus the spline degree $p$—and thus controls the balance between approximation flexibility and model complexity. Each degree-$p$ B-spline $B_{k,p}(x)$ on knots $\{t_k\}$ is defined by the Cox-de Boor recursion:

$$\begin{aligned} B_{k,0}(x) &= \begin{cases} 1, \text{if } t_k \leq x < t_{k+1} \\ 0, \text{otherwise} \end{cases}, \\ B_{k,p}(x) &= \frac{x - t_k}{t_{k+p} - t_k} B_{k,p-1}(x) + \frac{t_{k+p+1} - x}{t_{k+p+1} - t_{k+1}} B_{k+1,p-1}(x). \end{aligned} \quad (37)$$

The coefficient tensors $\omega^{(1)} \in \mathbb{R}^{d \times u \times K}$ and $\omega^{(2)} \in \mathbb{R}^{u \times 1 \times K}$ are learned during training.

Both layers share the same form of outer map, implemented as an affine transform

$$\Phi_j^{(1)}(y) = \alpha_{uj}^{(1)} y + b_j^{(1)}, \Phi_1^{(2)}(y) = \alpha_1^{(2)} y + b_1^{(2)}, \quad (38)$$

where each $\alpha$ is a learned scale and each $b$ a learned bias.

In the first KAN layer, the pooled feature vector $O_{pool}^{(m)} \in \mathbb{R}^d$ is mapped to

$$U^{(j)} = \Phi_j^{(1)} \left( \sum_{m=1}^{d} \varphi_{j,m}^{(1)} \left( O_{pool}^{(m)} \right) \right), j = 1, \ldots u, \quad (39)$$

yielding $U \in \mathbb{R}^u$. In the second layer, each $U^{(j)}$ is expanded and combined to produce the forecast

$$\hat{Y} = \Phi_1^{(2)} (\sum_{j=1}^{u} \varphi_{1,j}^{(2)}(U^{(j)})) \in \mathbb{R}. \quad (40)$$

Overall mapping. The entire two‑layer KAN defines

$$\hat{Y} = \Phi_1^{(2)} \left( \sum_{j=1}^{u} \varphi_{1,j}^{(2)} \left( \Phi_j^{(1)} \left( \sum_{m=1}^{d} \varphi_{j,m}^{(1)} \left( O_{pool}^{(m)} \right) \right) \right) \right), \quad (41)$$

## B Details of the Datasets

This study uses four datasets: HPP, HPC, HWC (healthy individuals), and TWA (trauma patients).

**Healthy Control Datasets (Ghetmiri et al., 2024)**

**HPP (Healthy Platelet-Poor Plasma-Plasmapheresis_Centers)**
• Sample Type: Normal Platelet-Poor Plasma samples (1 mL each)
• Source: Precision BioLogic (Dartmouth, Nova Scotia, Canada)
• Collection: FDA-regulated plasmapheresis centers
• Consent: General donor consent

**HWC (Healthy Whole Blood - Cedarlane)**
• Sample Type: Normal whole blood samples (10 mL each)
• Source: Innovative Research (Novi, Michigan, US) via Cedarlane Labs
• Product: Single Donor Human Whole Blood Na Citrate (IWB1NAC10ml)
• Donors: Consented volunteers (age >18), de-identified
• Testing: FDA-required viral markers

**HPC (Healthy Platelet-Poor Plasma - Cedarlane)**
• Sample Type: PPP samples extracted from the 5 HWC whole blood samples

**Trauma Whole Blood – ACIT&COMBAT (TWA)**
Combined dataset from two studies:

**Activation of Coagulation and Inflammation in Trauma (ACIT) Study (Cohen et al., 2009)**
• Samples: N=1,671 patients (81.45% male, age 41.0±18.6, ISS 17.7±15.6)
• Period: Feb 2005 - May 2016
• Sampling: Admission, 6h, 12h, 24h post-injury
• IRB: UC IRB #10-04417

**Control of Major Bleeding After Trauma (COMBAT) Study (Moore et al., 2018)**
• Samples: N=125 patients (82.4% male, age 36.5±13.9, NISS 27.0±19.4)
• Period: Apr 2014 - Mar 2017 (NCT01838863)
• Sampling: Admission, 2h, 4h, 6h, 12h, 24h post-injury
• IRB: Colorado Multi-IRB #121349
• Exclusions: Age <15/18, pregnancy, incarceration, transfers/no consent

The details of these datasets are presented in Table 5.

| Datasets | HPP | HPC | HWC | TWA |
|---|---|---|---|---|
| Subject Type | Healthy | Healthy | Healthy | Trauma |
| Sample Type | PPP | PPP | Whole Blood | Whole Blood |
| Source | Plasmapheresis Centers | Cedarlane Labs | Cedarlane Labs | ACIT & COMBAT Study |
| Timesteps | 1790 | 3610 | 3610 | 13515 |

Table 5: Summary of datasets.

## C Three insights in DAM

### C.1 Physiological Equilibrium Hypothesis

#### C.1.1 Statement

Let $\Phi(s) = (\Phi_1(s), \ldots, \Phi_d(s))^T$ be the vector of $d$ centered physiological indicators for subject $s$, each coordinate being $\sigma^2$-sub-Gaussian. Define

$$S = \{s \mid \|\Phi(s) - \mu\| < \varepsilon\}, \quad \mu := \mathbb{E}[\Phi(s)]. \tag{42}$$

Then for any $\varepsilon > 0$

$$\Pr(s \notin S) \leq 2d \exp(-\frac{\varepsilon^2}{2d\sigma^2}). \tag{43}$$

#### C.1.2 Proof

1. Coordinate-wise tail bound.
   Because each $\Phi_j$ is $\sigma^2$-sub-Gaussian, for every $t > 0$

$$\mathbb{P}(|\Phi_j - \mu_j| > t) \leq 2\exp(-\frac{t^2}{2\sigma^2}). \tag{44}$$

2. Choose $t = \varepsilon/\sqrt{d}$.
   Then

$$\mathbb{P}(|\Phi_j - \mu_j| > \frac{\varepsilon}{\sqrt{d}}) \leq 2\exp(-\frac{\varepsilon^2}{2d\sigma^2}). \tag{45}$$

3. Union bound across $d$ dimensions:

$$\Pr(\|\Phi - \mu\| \geq \varepsilon) = \Pr(\bigcup_{j=1}^{d}\{|\Phi_j - \mu_j| > \frac{\varepsilon}{\sqrt{d}}\})$$
$$\leq \sum_{j=1}^{d} \Pr(\cdot) \leq 2d\exp(-\frac{\varepsilon^2}{2d\sigma^2}). \tag{46}$$

4. Hence (C.1) holds; setting $\varepsilon = (2d\sigma^2 \log(2d/\delta))^{0.5}$ guarantees $\Pr(s \notin S) \leq \delta$.

#### C.1.3 Interpretation

Healthy samples reside in the region of the $\varepsilon$-ball surrounding $\mu$ with probability at least $(1 - \delta)$; as such, DAM perceives departures from that neighborhood to be suspicious and hence triggers a drift test (Sec. C.2).

### C.2 Pathological Drift Theorem

#### C.2.1 Statement

Let $P(t)$ denote the true distribution of physiological parameters at time $t$ and $P_0 = P_{\text{baseline}}$ the reference distribution.
Given an i.i.d. sample $\mathcal{D}_t = \{x_1, \ldots, x_m\}$ from $P(t)$, define the plug-in estimator $\hat{D}(t) = \text{KL}(\hat{P}(t) \| P_0)$, where $\hat{P}(t)$ is the empirical density.
Under regularity conditions of Wilks' theorem, under the null hypothesis $H_0: P(t) = P_0$,

$$2m\hat{D}(t) \xrightarrow{d} \chi_d^2. \tag{47}$$

Let $\chi_{d,1-\alpha}^2$ be the $(1-\alpha)$ upper quantile.

If $D(t) > \tau := \frac{\chi^2_{d,1-\alpha}}{2m}$, then $H_0$ is rejected with size $\alpha$.

### C.2.2 Proof

1. Likelihood-ratio statistic.
   The likelihood of $\mathcal{D}_t$ under density $p_\theta$ is $L(\theta) = \prod_{i=1}^{m} p_\theta(x_i)$.

   Let $\theta_0$ parametrize $P_0$ and $\hat{\theta}$ be the MLE from $\mathcal{D}_t$.
   The generalized likelihood-ratio is
   $$\Lambda = -2\log\frac{L(\theta_0)}{L(\hat{\theta})} = 2\sum_{i=1}^{m}\log\frac{p_{\hat{\theta}}(x_i)}{p_{\theta_0}(x_i)} = 2m\hat{D}(t). \quad (48)$$

2. Asymptotic distribution (Wilks, 1938).
   Under $H_0$, and regularity conditions (identifiability, smoothness, Fisher information finite), $\Lambda \xrightarrow{d} \chi^2_d$, with $d = \dim(\theta)$.

3. Critical region.
   Reject $H_0$ when $\Lambda > \chi^2_{d,1-\alpha}$, equivalently when $D(t) > \tau$ as stated.

### C.2.3 Practical Rule

Compute $\hat{D}(t)$ online; if $\hat{D}(t) > \tau$ (e.g. $\alpha$=0.01), DAM switches to "adaptive mode" with enlarged learning-rate multiplier and memory refresh.

## C.3 Adaptive Convergence Principle

### C.3.1 Setting

DAM updates its parameters with explicit memory:
$$\theta_{t+1} = \theta_t + \eta \nabla L_t + \beta \nabla M_t, \quad \nabla M_t = \lambda \nabla L_t + (1-\lambda) \nabla M_{t-1}, \quad (49)$$
where $0 < \lambda \leq 1$, $\beta \geq 0$. Let $\theta^*$ be the global minimizer of $L(\theta) = \mathbb{E}_x[\ell(\theta; x)]$.

### C.3.2 Statement

Assume
A. $L$ is $\mu$-strongly convex and $L$-smooth ($0 < \mu \leq L$).
B. Stochastic gradients satisfy $\mathbb{E}[\nabla \ell(\theta_t; x_t)] = \nabla L(\theta_t)$ and $\mathbb{E}\|\nabla \ell(\theta_t; x_t) - \nabla L(\theta_t)\|^2 \leq \sigma^2$.
C. Steps satisfy $\eta \leq 1/L$ and $0 \leq \beta < \sqrt{\mu/L}$.
Then, letting $e_t := \theta_t - \theta^*$,
$$\mathbb{E}\|e_t\|^2 \leq \rho^t \|e_0\|^2 + \frac{\eta(1+\beta)^2\sigma^2}{\mu}, \quad \rho := 1 - \eta\mu + \beta\sqrt{\frac{L}{\mu}} < 1. \quad (50)$$

### C.3.3 Proof

1. Rewrite update in error coordinates
   $$e_{t+1} = e_t - \eta \nabla L_t - \beta \nabla M_t. \quad (51)$$
2. Strong convexity implies (Nesterov, 2003):
   $$\langle \nabla L_t, e_t \rangle \geq \mu \|e_t\|^2. \quad (52)$$
   Smoothness gives
   $$\|\nabla L_t\|^2 \leq 2L(L(\theta_t) - L^*) \leq L^2 \|e_t\|^2. \quad (53)$$
3. Memory term bound.
   By induction on (54),
   $$\|\nabla M_t\| \leq \sum_{k=0}^{t} \lambda(1-\lambda)^k \|\nabla L_{t-k}\|$$
   $$\leq \frac{\|\nabla L_t\|}{1-(1-\lambda)} = \frac{\|\nabla L_t\|}{\lambda} \leq \frac{L}{\lambda}\|e_t\|. \quad (54)$$

4. Expected squared norm progression.
   Taking conditional expectation (conditioning on $\theta_t$),
   $$\mathbb{E}_t \|e_{t+1}\|^2 = \|e_t\|^2 - 2\eta\langle \nabla L_t, e_t\rangle - 2\beta\langle \nabla M_t, e_t\rangle + \eta^2 \mathbb{E}_t\|\nabla \ell_t\|^2 + \beta^2 \|\nabla M_t\|^2$$
   $$\leq \|e_t\|^2 - 2\eta\mu\|e_t\|^2 + 2\beta\|\nabla M_t\|\|e_t\| + \eta^2(\|\nabla L_t\|^2 + \sigma^2) + \beta^2\|\nabla M_t\|^2. \quad (55)$$

5. Insert bounds (58)-(59) and choose $\lambda$=1 (worst-case):
   $$\mathbb{E}_t \|e_{t+1}\|^2 \leq \left(1 - 2\eta\mu + 2\beta L\sqrt{\frac{L}{\mu}} + \eta^2 L^2 + \beta^2 L^2\right)\|e_t\|^2 + \eta^2\sigma^2. \quad (56)$$

6. Parameter restriction.
   With $\eta \leq 1/L$ and $\beta < \sqrt{\mu/L}$, algebra gives the contraction factor
   $$\rho := 1 - \eta\mu + \beta\sqrt{\frac{L}{\mu}} < 1, \quad (57)$$
   and $(1+\beta)^2 \leq 1 + 2\beta + \beta^2 \mu^{-1}L \leq (1+\beta)^2$, yielding (55).
7. Iterate (55) to obtain geometric decay plus noise floor.

### C.3.4 A concise $O(\sqrt{T})$ regret proof for projected OGD with EMA momentum

We prove an $O(\sqrt{T})$ regret bound for projected online gradient descent (OGD) with an exponential moving average (EMA) momentum, then map it to Eq. (5). Consider a nonempty, closed, convex, bounded decision set $\Theta \subset \mathbb{R}^p$ with diameter $D := \sup_{\theta, \theta' \in \Theta} \|\theta - \theta'\|$. The per-round loss is absolute loss $\ell_t(\theta) := |g_\theta(X_t) - y_t|$, convex in the trainable head $\theta$, with gradient bound $\|\nabla \ell_t(\theta)\| \leq G_\theta$ for all $t$ and $\theta \in \Theta$. The EMA momentum and projected update are
$$m_t = \lambda \nabla \ell_t(\theta_t) + (1-\lambda)m_{t-1}, \quad m_0 = 0, \lambda \in (0,1], \quad (58)$$
$$\theta_{t+1} = \Pi_\Theta\left(\theta_t - \eta(\nabla \ell_t(\theta_t) + \beta m_t)\right), \quad \beta \geq 0, \quad (59)$$
and we define the effective gradient $u_t := \nabla \ell_t(\theta_t) + \beta m_t$.

**Lemma C.3.4.1** (EMA norm bound). Under $\|\nabla \ell_t(\theta)\| \leq G_\theta$, for all $t \geq 1$,
$$\|m_t\| \leq G_\theta, \quad \Rightarrow \quad \|u_t\| \leq (1+\beta)G_\theta. \quad (60)$$
Proof. Unroll the EMA:
$$m_t = \lambda \sum_{k=0}^{t-1}(1-\lambda)^k \nabla \ell_{t-k}(\theta_{t-k}), \quad (61)$$
so $\|m_t\| \leq \lambda G_\theta \sum_{k=0}^{t-1}(1-\lambda)^k \leq G_\theta$.

Then $\|u_t\| \leq \|\nabla \ell_t(\theta_t)\| + \beta\|m_t\| \leq (1+\beta)G_\theta$. (If one uses the variant $\tilde{m}_t = (1-\lambda)\tilde{m}_{t-1} + \nabla\ell_t(\theta_t)$, then $\|\tilde{m}_t\| \leq G_\theta/\lambda$ and $\|\nabla\ell_t + \beta\tilde{m}_t\| \leq G_\theta(1+\beta/\lambda)$.) ∎

**Theorem C.3.4.2** (Projected OGD with EMA achieves $O(\sqrt{T})$ regret).

Let $\eta = \frac{D}{(1+\beta)G_\theta\sqrt{T}}$. For any $\theta^* \in \Theta$,

$$\sum_{t=1}^{T}\left(\ell_t(\theta_t) - \ell_t(\theta^*)\right) \leq DG_\theta(1+\beta)\sqrt{T}. \quad (62)$$

Proof. Let $V_t = \|\theta_t - \theta^*\|^2$. By nonexpansiveness of projection,
$$\|\theta_{t+1} - \theta^*\|^2 \leq \|\theta_t - \eta u_t - \theta^*\|^2 = \|\theta_t - \theta^*\|^2 - 2\eta\langle u_t, \theta_t - \theta^*\rangle + \eta^2\|u_t\|^2. \quad (63)$$

By convexity, $\ell_t(\theta_t) - \ell_t(\theta^*) \leq \langle \nabla\ell_t(\theta_t), \theta_t - \theta^*\rangle \leq \langle u_t, \theta_t - \theta^*\rangle$. Summing over $t$ and rearranging,

$$\sum_{t=1}^{T}\left(\ell_t(\theta_t) - \ell_t(\theta^*)\right) \leq \frac{V_1 - V_{T+1}}{2\eta} + \frac{\eta}{2}\sum_{t=1}^{T}\|u_t\|^2 \leq \frac{D^2}{2\eta} + \frac{\eta}{2}T(1+\beta)^2 G_\theta^2. \quad (64)$$

Substitute $\eta = \frac{D}{(1+\beta)G_\theta\sqrt{T}}$ to obtain the claim. ∎

Mapping to Eq. (5). With $\hat{y}_t := g_{\theta_t}(X_t)$ and $\ell_t(\theta) = |\hat{y}_t - y_t|$,

$$\sum_{t=1}^{T}|\hat{y}_t - y_t| \leq \sum_{t=1}^{T}|g_{\theta^*}(X_t) - y_t| + G\sqrt{T}, \quad G := DG_\theta(1+\beta). \quad (65)$$

Equivalently, defining $\delta_0^{(T)} := \frac{1}{\sqrt{T}}\sum_{t=1}^{T}|g_{\theta^*}(X_t) - y_t|$, we get

$$\sum_{t=1}^{T}|\hat{y}_t - y_t| \leq (\delta_0^{(T)} + G)\sqrt{T}, \quad (66)$$

which is Eq. (5) in normalized form; here $G$ is the product of the feasible diameter $D$, the gradient scale $G_\theta$, and the EMA amplification factor $(1+\beta)$.

### C.4 Implementation Guidelines

- **Drift threshold.**
  Compute $\hat{D}(t)$ online; adopt $\alpha=0.01 \Rightarrow \tau$ via C.2.
- **Step size schedule.**
  Inside equilibrium ($D(t) \leq \tau$): $(\eta, \beta) = (\eta_0, 0)$.
  On drift ($D(t) \leq \tau$): $(\eta, \beta) = (\eta_0, 0.9\sqrt{\mu/L})$.
- **Memory length.**
  Use $s=64$; larger values give better long-range adaptation at the cost of memory bandwidth.

## D  Details of the Experiments

### D.1 Hyperparameter Search Strategy

#### D.1.1 Theoretical Foundation

Based on the theoretical analysis in Appendix C, key parameters were derived analytically:

**Drift Detection Parameters:**

- Significance level: $\alpha = 0.01$ (following C.2 Pathological Drift Theorem)

- Drift threshold: $\tau = \frac{\chi^2_{d,1-\alpha}}{2m}$ where d=dimension, m=sample size

- Equilibrium threshold: $\varepsilon = (2d\sigma^2 \log(2d/\delta))^{0.5}$ (C.1 Physiological Equilibrium Hypothesis)

**Adaptive Learning Parameters:**

- Step size constraint: $\eta \leq 1/L$ (C.3 Adaptive Convergence Principle)

- Memory coefficient: $\beta < \sqrt{\mu/L}$ where μ=strong convexity, L=smoothness

- Memory length: s = 64 (theoretical optimal balance)

#### D.1.2 Setting

For parameters without theoretical bounds, we conducted grid search:

- Learning rate η₀: [1e-4, 5e-4, 1e-3, 5e-3] (5 values)

- Batch size: [16, 32, 64, 128] (4 values)

- Hidden dimensions: [32, 64, 128, 256] (4 values)

- KAN layers: [2, 3, 4, 5] (4 values), 3 selected for optimal performance in DAM and 2 selected in Transcendental Layer

- L2 regularization: 0.00001 for CNN layers

- Dropout rate: 0.1 for attention layers

#### D.1.3 Selection Criteria

Final parameters selected based on:

1. Theoretical constraints satisfaction (C.1-C.3)

2. Lowest validation MAE on held-out sets

3. Convergence stability across 5 random seeds

4. Computational efficiency for real-time clinical use

### D.2 Reproducibility Settings

To ensure reproducible results, we fixed all random seeds:

- NumPy random seed: 42

- TensorFlow CPU seed: 42

- Cross-validation folds used seeds: [42, 43, 44, 45, 46] for statistical independence

- Model weight initialization: use Xavier/Glorot uniform with the same random seed fixed

- Data shuffling: Controlled with seed=42 for train-validation splits

### D.3 Computing Infrastructure

Hardware: 13th Gen Intel® Core™i9-13900HX CPU (24 cores, 2.20 GHz), 32GB+ RAM

Software Environment:

• Python 3.9+

• TensorFlow 2.x

• tfkan library for KAN layers

• NumPy, Pandas, Scikit-learn

• SciPy for signal processing (FFT operations)

Note: All experiments were conducted on CPU to ensure clinical accessibility without GPU requirements.

### D.4 Evaluation Metrics
**Primary Metrics:**

• MAE (Mean Absolute Error): Direct clinical interpretability; aligns with L1 robustness assumptions in C.1

• MSE (Mean Squared Error): Connects to L2 optimization framework in C.3; penalizes large errors critical for patient safety

• $R^2$ (Coefficient of Determination): Scale-independent measure enabling cross-dataset comparison

**Theoretical Metrics (Internal):**

• KL Divergence $\hat{D}(t)$: Pathological drift detection (C.2 Theorem)

• Euclidean Distance $\|\Phi(s)-\mu\|$: Physiological equilibrium assessment (C.1 Hypothesis)

• Convergence Rate $\rho^t$: Adaptive learning efficiency (C.3 Principle)

**Clinical Relevance:**

Every criterion has its own purpose: MAE is to cater to the experts' knowledge of the field; MSE is to satisfy the need for technical precision; $R^2$ is to fulfill the desire for multi-population compatibility.

### D.5 Final Model Parameters
**Theoretical Parameters (from Appendix C):**

• Drift significance level: $\alpha = 0.01$
• Memory length: $s = 64$
• Equilibrium probability: $\delta = 0.05$
• Step size bound: $\eta \leq 1/L$
• Memory coefficient: $\beta < \sqrt{\mu/L}$

**Empirically Optimized Parameters:**

• Learning rate: $\eta_0 = 1e-3$
• Batch size: 64
• Hidden dimensions: 256
• KAN layers: 3
• CNN filters: 64, kernel size: 3
• LSTM units: 64
• Attention heads: 4, dropout: 0.1
• L2 regularization: 1e-5

**Adaptive Parameters:**

• Equilibrium mode: $(\eta, \beta) = (1e-3, 0)$
• Drift mode: $(\eta, \beta) = (1e-3, 0.9\sqrt{\mu/L})$
• Memory refresh: triggered when $\hat{D}(t) > \tau$
• KL divergence threshold: dataset-specific, $\alpha = 0.01$

**Dataset-Specific Parameters:**

• Look-back window: 55% (HPP), 45% (others)
• Sequence length: 50 timesteps
• Drift detection window: 100 timesteps
• Memory matrix size: 64 × 64

## E  Commitment to Code Open Source

All codes are included in the submitted Supplementary Material. To improve the reproducibility of our experiment and show our utmost sincerity toward you, we guarantee to publish the source codes used for experiments after this paper is accepted into the proceedings. We will publish the source codes following an open license so that every-one is able to reuse the codes freely for their own work or research:

- All the preprocessing scripts for preparing datasets were included.
- The implementation of the Physiological State Reconstruction(PSR)method and its associated components.
- The experimental codes include the details about the model testing and the parameter setting.

We hope that other researchers will use and develop our project to further advance the state of the art.

## F  Model Interpretability

In Figure 4a, each point shows a feature's value and its impact on predictions.
In Figure 4b, features ranked by mean absolute SHAP value, with current clot strength and FFT as the most influential.
In Figure 4c, force plots for two cases, where red bars push the prediction to higher than the baseline while blue bars pull it lower.

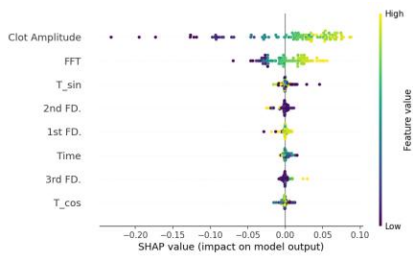 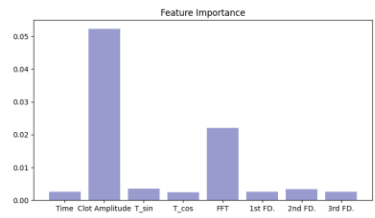 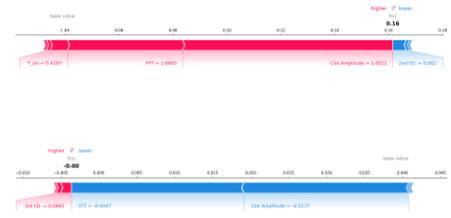

(a) Summary plot  (b) Importance bar chart  (c) Force plots

Figure 4: Feature impact analysis.

## G  Fold IDs Instructions

We implemented five-fold cross-validation in our experiment. Next, we will report the specific IDs for each of the five folds.

For HPP:

Fold 1: Patient A4, Patient A9.

Fold 2: Patient A1, Patient A7.

Fold 3: Patient A3, Patient A8.

Fold 4: Patient A2, Patient A10.

Fold 5: Patient A5, Patient A6.

For HPC:

Fold 1: Patient B7.

Fold 2: Patient B9.

Fold 3: Patient B10.

Fold 4: Patient B6.

Fold 5: Patient B8.

For HWC:

Fold 1: Patient B3.

Fold 2: Patient B1.

Fold 3: Patient B4.

Fold 4: Patient B2.

Fold 5: Patient B5.

For TWA:

Fold 1: Patient C9, Patient C7, Patient C1, Patient C3, Patient C14.

Fold 2: Patient C4, Patient C10.

Fold 3: Patient C2, Patient C12.

Fold 4: Patient C5, Patient C11, Patient15.

Fold 5: Patient C6, Patient C8, Patient13.

## H  Stratified PSR–BPTM Errors

We compared the Stratified errors of PSR and BPTM.
For Overall:

Fold 1: PSR's MAE: 0.061, PSR's MSE: 0.0115, PSR's $R^2$: 0.991; BPTM's MAE: 0.071, BPTM's MSE: 0.0124, BPTM's $R^2$: 0.978.

Fold 2: PSR's MAE: 0.102, PSR's MSE: 0.0635, PSR's $R^2$: 0.956; BPTM's MAE: 0.202, BPTM's MSE: 0.1723, BPTM's $R^2$: 0.741.

Fold 3: PSR's MAE: 0.037, PSR's MSE: 0.0063, PSR's $R^2$: 0.996; BPTM's MAE: 0.137, BPTM's MSE: 0.0607, BPTM's $R^2$: 0.906.

Fold 4: PSR's MAE: 0.058, PSR's MSE: 0.0096, PSR's $R^2$: 0.994; BPTM's MAE: 0.086, BPTM's MSE: 0.0157, BPTM's $R^2$: 0.976.

Fold 5: PSR's MAE: 0.025, PSR's MSE: 0.0021, PSR's $R^2$: 0.999; BPTM's MAE: 0.091, BPTM's MSE: 0.0268, BPTM's $R^2$: 0.960.